\documentclass{article} 
\usepackage{iclr2025_conference,times}

\usepackage{amsmath,amsfonts,bm}

\def\eqref#1{equation~\ref{#1}}

\def\1{\bm{1}}

\DeclareMathAlphabet{\mathsfit}{\encodingdefault}{\sfdefault}{m}{sl}
\SetMathAlphabet{\mathsfit}{bold}{\encodingdefault}{\sfdefault}{bx}{n}

\usepackage{hyperref}
\usepackage{url}

\usepackage{graphicx}
\usepackage{xspace}
\usepackage{booktabs}
\usepackage{adjustbox}
\usepackage{amssymb}
\usepackage{colortbl}
\usepackage{xcolor}
\usepackage{longtable}

\usepackage[most]{tcolorbox}
\usepackage{colortbl}
\usepackage{cleveref}
\usepackage{arydshln}
\usepackage{amsthm}
\usepackage{wrapfig}
\usepackage{caption}
\usepackage{tabularx}
\usepackage{pifont}
\usepackage{subcaption}
\usepackage{fontawesome5}

\newcommand{\grayrow}{\rowcolor[gray]{0.95}}

\newcommand{\eg}{e.g.,\xspace}
\newcommand{\ie}{i.e.,\xspace}

\newcommand{\tabincell}[2]{\begin{tabular}{@{}#1@{}}#2\end{tabular}}

\theoremstyle{definition}
\newtheorem{definition}{Definition}[section]

\tcbuselibrary{theorems}

\newtcbtheorem{hypothesis}{Hypothesis}{
  colback=yellow!5,
  colframe=yellow!50!black,
  fonttitle=\bfseries,
  separator sign={.},
  description delimiters={}{},
}{hypo}

\newcommand{\downperf}{\textcolor{red}{$\blacktriangledown$}\xspace}
\newcommand{\upperf}{\textcolor{teal}{$\blacktriangle$}\xspace}
\makeatletter
\newcommand{\github}[1]{%
   \href{#1}{\faGithubSquare}%
}
\makeatother

\title{Intriguing Properties of Large Language and Vision Models}

\iclrfinalcopy

\author{
Young-Jun Lee \textsuperscript{\rm 1} \quad
Byungsoo Ko \textsuperscript{\rm 1} \quad
Han-Gyu Kim \textsuperscript{\rm 2} \quad
Yechan Hwang \textsuperscript{\rm 1} \quad 
Ho-Jin Choi \textsuperscript{\rm 1}
\\
\textsuperscript{\rm 1} School of Computing, KAIST \quad
\textsuperscript{\rm 2} NAVER Cloud Multimodal AI
\\
\texttt{\{yj2961, yemintmint, hojinc\}@kaist.ac.kr} \\
\texttt{Kobiso62@gmail.com} \quad \texttt{hangyu.kim@navercorp.com} \\
{\github{} Code:~\texttt{\url{https://github.com/passing2961/IP-LLVM}}} 
}

\begin{document}

\maketitle

\begin{abstract}
    Recently, large language and vision models (LLVMs) have received significant attention and development efforts due to their remarkable generalization performance across a wide range of tasks requiring perception and cognitive abilities. A key factor behind their success is their simple architecture, which consists of a vision encoder, a projector, and a large language model (LLM). Despite their achievements in advanced reasoning tasks, their performance on fundamental perception-related tasks (\eg MMVP) remains surprisingly low. This discrepancy raises the question of how LLVMs truly perceive images and exploit the advantages of the vision encoder. To address this, we systematically investigate this question regarding several aspects: \textit{permutation invariance}, \textit{robustness}, \textit{math reasoning}, \textit{alignment preserving} and \textit{importance}, by evaluating the most common LLVM's families (\ie LLaVA) across 10 evaluation benchmarks. Our extensive experiments reveal several intriguing properties of current LLVMs: (1) they internally process the image in a global manner, even when the order of visual patch sequences is randomly permuted; (2) they are sometimes able to solve math problems without fully perceiving detailed numerical information; (3) the cross-modal alignment is overfitted to complex reasoning tasks, thereby, causing them to lose some of the original perceptual capabilities of their vision encoder; (4) the representation space in the lower layers ($<25\%$) plays a crucial role in determining performance and enhancing visual understanding. Lastly, based on the above observations, we suggest potential future directions for building better LLVMs and constructing more challenging evaluation benchmarks.
\end{abstract}

\section{Introduction} \label{main_sec:intro}

Large Language and Vision Models (LLVMs)\citep{liu2024visual, liu2024llava, lee2024meteor, lee2024trol} have demonstrated remarkable generalization capabilities across a wide variety of tasks, including coding and mathematics, showcasing their potential for practical applications. These impressive advancements have been achieved through a straightforward yet effective architecture based on the concept of \textit{model-stitching}\citep{lenc2015understanding, bansal2021revisiting}. This approach integrates a pre-trained vision encoder~\citep{radford2021learning} with a pre-trained large language model (LLM)~\citep{touvron2023llama, zheng2023judging} via a simple cross-modal alignment module. This method significantly benefits from the power of well-established pre-trained representations. Consequently, this structure has become the \textit{de facto} standard in the field, extending into other modality domains such as video, audio, and unified modalities~\citep{xie2024show,sharegpt4o}.

Despite their significant generalization performance, recent studies have revealed several interesting phenomena regarding LLVMs. For instance, they struggle with tasks that are easy for humans to perceive (\eg MMVP~\citep{tong2024eyes}, BLINK~\citep{fu2024blink}) and have limited understanding of domain-specific images~\citep{zhai2024investigating,verma2024cross}. In contrast to the computer vision domain, where demystifying the properties of vision encoders has been more thoroughly explored~\citep{naseer2021intriguing,kim2023demystifying,vishniakov2023convnet}, the underlying properties of LLVMs are still largely under-explored. Therefore, in this work, we scrutinize the current \textit{de facto} structure of LLVMs under various partial conditions, such as \textit{permutation}, \textit{occlusion}, and \textit{synthetic images}.

In this paper, we systematically conduct a comprehensive evaluation of widely used LLVM families, specifically the \texttt{LLaVA}-series, across 10 diverse benchmarks, which encompass tasks such as math, chart, and basic perception. Our extensive experiments reveal several intriguing properties of current LLVMs, which we summarize as follows:
\begin{itemize}
    \item In LLVMs, the visual patch tokens processed through the projector exhibit varying magnitudes of localized visual information. Remarkably, even when the order of these patch sequences is randomly shuffled before being fed into the LLM, the performance does not significantly degrade. For instance, in the case of \texttt{LLaVA 1.5}~\citep{li2024llava}, the average performance drop across 10 benchmarks is 0.19 ($< 1\%$), indicating that LLVMs exhibit permutation-invariant properties.
    
    \item LLVMs effectively handle tasks when given synthetic versions of the MathVista~\citep{lu2023mathvista} dataset, with only a small performance decline (1.8$\%$ for \texttt{LLaVA 1.5}). Furthermore, we discovered that, in certain scenarios, LLVMs can solve problems even without access to the full image, including detailed numerical and chart elements.

    \item Following alignment and visual instruction tuning, LLVMs fail to preserve their initial perceptual capacities, with up to a 20\% drop in image classification tasks (\eg CIFAR-100~\citep{krizhevsky2009learning}), a phenomenon known as catastrophic forgetting~\citep{zhai2024investigating}. Furthermore, they struggle to understand shared-world concepts within the representation space, according to the platonic representation hypothesis~\citep{huh2024platonic}.

    \item Our analysis of model behavior reveals that LLVMs tend to concentrate more on the central region of the image. Furthermore, the lower layers in LLVM architectures are crucial for better generalization. In these layers (\ie the bottom 20\% of the LLM layers), the model primarily processes visual information, while the higher layers focus on interpreting the text.

\end{itemize}

In addition to our findings, we present and discuss several points regarding LLMs and evaluation benchmarks. Specifically, we highlight the need to develop more interactive and complex evaluation benchmarks to mitigate selection bias~\cite{zheng2023large} and improve applicability to real-world scenarios. Furthermore, when developing new LLMs, it is crucial to preserve cross-modal alignment. We hope that our findings will assist other ML researchers and engineers in building a new paradigm for LLMs.

\section{Related Works} \label{main_sec:related_work}

\paragraph{Model-Stitching.} The model-stitching~\citep{lenc2015understanding,bansal2021revisiting} is a technique first introduced to study the internal representations of neural networks by measuring the representational similarity between two given models. Consider two models defined as $f = f^m \circ \cdots \circ f^1$ and $g = g^n \circ \cdots \circ g^1$. Specifically, the \textit{stitched} model is formalized as $\mathcal{F} = g^{n} \circ \cdots \circ g^{k+1} \circ s \circ f^{k} \circ \cdots \circ f^1$, where $s$ is a simple stitching layer (e.g., a linear layer or a $1 \times 1$ convolution). Therefore, even if the two models $f$ and $g$ differ in training methodology (e.g., supervised vs.\ self-supervised) or modalities (e.g., text vs.\ image), if $\mathcal{F}$ exhibits good performance, then $f$ and $g$ have strongly correlated and compatible internal representations at layer $k$, apart from the stitching layer $s$. \citet{merullo2022linearly} have the similar concept of \textit{model-stitching} to verify their strong hypothesis that the conceptual representations from a frozen LLM and a visual encoder are sufficiently similar such that a simple linear mapping layer can align them.

\paragraph{Large Language and Vision Models.} Recent advancements in LLVMs have predominantly adopted simplistic yet highly effective architectures, notably through the model-stitching concept. Numerous prior studies have introduced various design modifications to bridge the performance gap with closed-source LLVMs~\citep{gpttechnical,claude3series2024}. These efforts include focusing intently on high-resolution processing~\citep{li2024mini,liu2024llava,shi2024eagle}, implementing locality-enhanced projectors~\citep{cha2024honeybee}, and incorporating knowledge embeddings~\citep{lee2024meteor}, layer traversal technique~\citep{lee2024trol} and leveraging a diverse array of vision encoders~\citep{lu2024deepseek,tong2024eyes} have also been explored. Additionally, integrating external, task-specific computer vision modules~\citep{lee2024collavo,lee2024moai,jiao2024enhancing,lai2024lisa} and incorporating different modalities — including video and audio~\citep{wang2024qwen2,li2024llavaone,sharegpt4o,xie2024show} — have expanded the models' capabilities. Moreover, enabling the handling of interleaved input formats~\citep{li2024llava,xue2024xgen} has further broadened the versatility of these models. While these models have been developed based on a simplistic structure, \textit{model-stitching}, the effectiveness of this architecture remains under-explored.

\paragraph{Investigating Intriguing Properties of LLVMs.} Alongside these advancements, recent studies have investigated and uncovered several crucial properties of current LLVMs. For instance, some studies have rigorously evaluated LLVMs on basic perception tasks that are trivially easy for humans by introducing ``blind'' pairs of image datasets~\citep{tong2024eyes,fu2024blink,rahmanzadehgervi2024vision}. Other studies have explored cross-modal alignment by focusing on domain-specific visual capabilities~\citep{verma2024cross} and examining the alignment of representation spaces across modalities between independently pre-trained LLMs and vision encoders~\citep{li2024visionlanguagemodelsshare,huh2024platonic}. \citet{zhai2024investigating} examine the phenomenon of catastrophic forgetting in LLVMs within the context of image classification tasks. Additional studies~\citep{zhou2023analyzing,chen2024multi} analyze the persistent issue of object hallucination in LLVMs. Moreover, research has explored spatial reasoning capabilities~\citep{kamath2023s}. While vision encoders (\eg ViT~\citep{dosovitskiy2020image}, DeiT~\citep{touvron2021training}) in the computer vision field have been rigorously examined across a wide range of image settings, the study of these intriguing properties in LLVMs remains relatively under-explored. In this paper, we aim to address this by conducting an in-depth investigation into LLVMs, examining their permutation invariance, robustness, alignment preservation, and importance in scenarios involving occluded and synthesized images.

\section{Demystifying Intriguing Properties of LVLMs} \label{main_sec:intriguing}

In this section, we explore the intriguing properties of current LLVMs that have \textit{de facto} structure of \textit{modal-stitching} in terms of various aspects: permutation invariance, robustness to occlusion, synthetic data, alignment preserving, and importance.

\subsection{Background} \label{main_sec:background}

\paragraph{Overview of LVLM.} Current LVLMs $\mathcal{M}$ have widely adopted the \textit{model-stitching} architecture, which consists of three main components: a pre-trained vision encoder $f_\text{V}$, a projector $f_\text{P}$, and a pre-trained LLM $f_\text{L}$. The overall model is represented as $\mathcal{M} = f_\text{L} \circ f_\text{P} \circ f_\text{V}$. The vision encoder $f_\text{V}$ converts the input image $I \in \mathbb{R}^{3 \times H \times W}$ into visual features $\mathcal{F}_v \in \mathbb{R}^{N \times d_v}$, where $N = HW / P^2$ is the number of visual features, $P$ is the patch size, and $d_v$ is the dimension of the vision encoder's output. The projector $f_\text{P}$ transforms these visual features $\mathcal{F}_v$ into visual patch tokens $\mathbf{X}_\text{V} \in \mathbb{R}^{N \times d_l}$ in the representation space of the LLM, where $d_l$ is the embedding dimension of the LLM. This mapping allows the LLM to perceive and conceptually understand the given image. The LLM $f_\text{L}$ produces an appropriate response $\mathbf{Y} = \{y_i\}_{i=1}^{L_\mathbf{Y}}$ in an autoregressive manner, given both the visual patch tokens $\mathbf{X}_\text{V}$ and the text tokens $\mathbf{X}_\text{T} \in \mathbb{R}^{L_\mathbf{T} \times d_\mathbf{l}}$, where $L_\mathbf{T}$ denotes the length of the input text sequence, and $L_\mathbf{Y}$ is the length of the output sequence. The probability of generating the response is given by:
\begin{equation} \label{eq:lvlm_ar}
    p(\mathbf{Y} \mid \mathbf{X}_\text{V}, \mathbf{X}_\text{T}) = \prod_{i=1}^{L_\mathbf{Y}} p(y_i \mid \mathbf{X}_\text{V}, \mathbf{X}_\text{T}, y_{<i}) 
\end{equation}

\subsection{Evaluation Setup}

\paragraph{Evaluation Benchmarks.} To ensure a comprehensive and rigorous evaluation, we employ 10 standard and widely adopted benchmarks: MMVP~\citep{tong2024eyes}, Q-Bench~\citep{wu2023q}, MME~\citep{fu2023mme}, MMStar~\citep{chen2024we}, MM-Vet~\citep{yu2023mm}, LLaVA-W~\citep{liu2024visual}, MathVista~\citep{lu2023mathvista}, SQA-IMG~\citep{lu2022learn}, ChartQA~\citep{masry2022chartqa}, and AI2D~\citep{kembhavi2016diagram}. Detailed descriptions of each dataset are provided in Appendix~\ref{supp_sec:benchmark_description}.

\paragraph{Evaluation Models.} Recently, a large number of LVLM models have been actively introduced, owing to their remarkable flexibility and versatility across multiple domains. Consequently, it is challenging and inefficient to conduct holistic evaluations on all LVLMs. Therefore, we select most standard LLVMs: \texttt{LLaVA-1.5-7B}~\citep{li2024llava}, \texttt{LLaVA-NeXT-7B}~\citep{liu2024llava}, and \texttt{LLaVA-OneVision-8B}~\citep{li2024llavaone}. For our customized experiments, before evaluating LVLMs under diverse settings (\eg occlusion), we first attempt to reproduce the baseline performance of LVLMs on 10 evaluation benchmarks. To do this, we implement our customized evaluation toolkits by referring to the code of \texttt{UniBench}~\footnote{\url{https://github.com/facebookresearch/unibench}}~\citep{al2024unibench}. Detailed descriptions of each model are provided in Appendix~\ref{supp_sec:lvlm_description}.

\subsection{Do LLVMs Perceive Images Globally?} \label{sec:rq1}

Current LVLMs commonly adopt ViT~\citep{dosovitskiy2020image}-based vision encoders, such as CLIP ViT~\citep{radford2021learning} and SigLIP~\citep{zhai2023sigmoid}, making their image perception dependent on these encoders. Specifically, ViT is designed to learn interactions across all image patches, providing properties~\citep{naseer2021intriguing,vishniakov2023convnet} such as \textit{permutation invariance} and \textit{robustness to occlusion}. This raises the question of whether these ViT properties might transfer to current LVLM models.

\begin{wrapfigure}{r}{0.4\linewidth}
    \centering
    \vspace{-3em}
    \includegraphics[width=\linewidth]{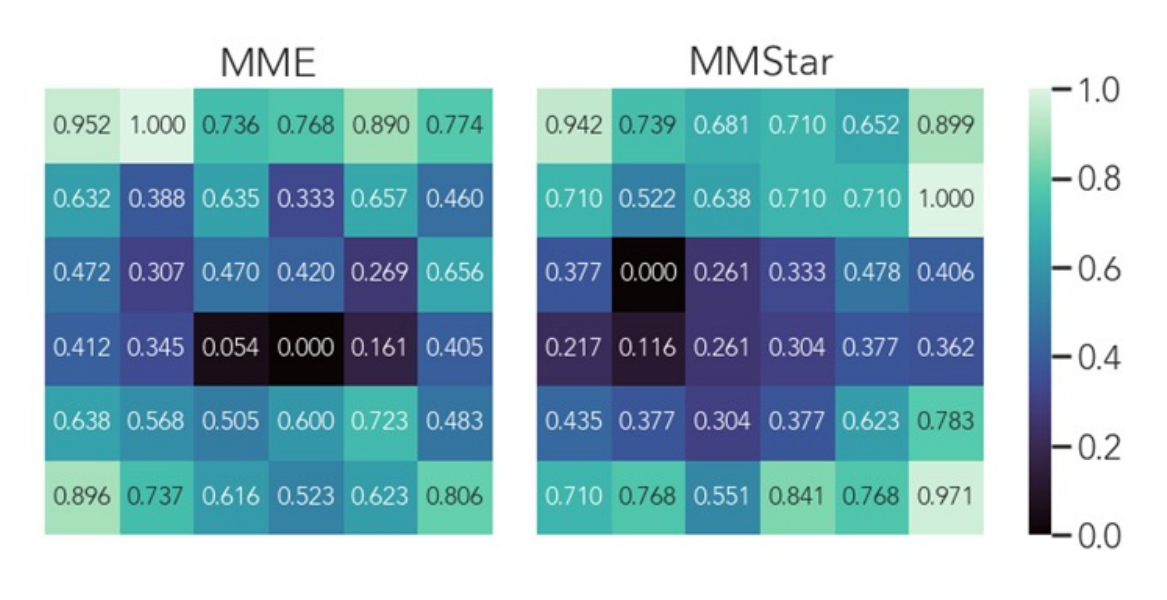}
    \vspace{-2em}
    \caption{
        We demonstrate the extent to which group-wise visual tokens capture region-specific information (\texttt{PIL}) for \texttt{LLaVA-1.5-7B} on the MMStar~\citep{chen2024we} and MME~\citep{fu2023mme}. Darker regions indicate areas where the model retains more localized information for those specific groups.
    }
    \vspace{-1em}
    \label{main_fig:vis_local_results}
\end{wrapfigure}

\paragraph{Each visual patch token encapsulates localized visual information.} We first investigate whether each visual patch token $\mathbf{X}_\text{V}$ from the projector $f_\text{P}$ captures a localized understanding of the patch area corresponding to its position in the image. Specifically, given an image $I$, the projector outputs $N$ visual patch tokens (\eg $N = 576$ for \texttt{LLaVA-1.5-7B}). We then select a single token (removing all others) and feed it into the LLM $f_\text{L}$. To quantify this, we define the patch information loss (\texttt{PIL}) as the ratio of the performance drop to the original performance. However, performing computations on each individual visual token is computationally intensive, especially for models such as \texttt{LLaVA 1.5-7B} that process 576 visual tokens arranged in a $24 \times 24$ grid of patches. To accelerate computation and reduce complexity, we aggregate the original $N$ visual tokens into $M$ tokens, where $M < N$, by grouping neighboring tokens. As shown in Figure~\ref{main_fig:vis_local_results}, the group-wise visual tokens in the \texttt{LLaVA-1.5-7B} model demonstrate varying levels of performance on the MMStar~\citep{chen2024we} and MME~\citep{fu2023mme}, suggesting that each token captures localized visual information rather than global concept understanding. Additionally, the central visual tokens contain more informative content compared to those at the edges.

{\renewcommand{\arraystretch}{1.35}
\begin{table}[!t]
\centering
\begin{adjustbox}{width=\linewidth}
\begin{tabular}{lccccccccccc}
\toprule
LLVMs & MMVP & Q-Bench & MME & MMStar & MM-Vet & LLaVA$^W$ & MathVista & SQA$^I$ & ChartQA & AI2D & Avg. $\Delta$\\
\midrule
LLaVA-1.5 & 34.67 & 59.73 & 1850.07 & 34.20 & 31.50 & 67.50 & 24.70 & 65.59 & 16.92 & 53.34 \\
\grayrow + \texttt{Perm.} & \tabincell{c}{36.00 \\ \small (\upperf 1.33)} & \tabincell{c}{59.60 \\ \small (\downperf 0.13)} & \tabincell{c}{1874.60 \\ \small (\upperf 24.53)} & \tabincell{c}{33.33 \\ \small (\downperf 0.87)} & \tabincell{c}{30.40 \\ \small (\downperf 1.10)} & \tabincell{c}{66.20 \\ \small (\downperf 1.30)} & \tabincell{c}{21.20 \\ \small (\downperf 3.50)} & \tabincell{c}{65.44 \\ \small (\downperf 0.15)} & \tabincell{c}{14.08 \\ \small (\downperf 2.84)} & \tabincell{c}{52.69 \\ \small (\downperf 0.65)} & \downperf 0.59 \\
LLaVA-NeXT & 36.67 & 63.55 & 1874.42 & 37.80 & 43.50 & 75.50 & 32.00 & 62.12 & 66.06 & 64.02 \\
\grayrow + \texttt{Perm.} & \tabincell{c}{37.33 \\ \small (\upperf 0.67)} & \tabincell{c}{62.54 \\ \small (\downperf 1.00)} & \tabincell{c}{1890.19 \\ \small (\upperf 15.78)} & \tabincell{c}{36.87 \\ \small (\downperf 0.93)} & \tabincell{c}{43.40 \\ \small (\downperf 0.10)} & \tabincell{c}{75.80 \\ \small (\upperf 0.30)} & \tabincell{c}{21.70 \\ \small (\downperf 10.30)} & \tabincell{c}{62.12 \\ \small (\downperf 0.00)} & \tabincell{c}{34.55 \\ \small (\downperf 31.51)} & \tabincell{c}{64.02 \\ \small (\downperf 0.00)} & \downperf 2.71 \\
LLaVA-OneVision & 60.67 & 77.26 & 1982.5 & 59.87 & 57.80 & 87.40 & 61.80 & 94.00 & 93.52 & 81.25 \\
\grayrow + \texttt{Perm.} & \tabincell{c}{59.33 \\ \small (\downperf 1.33)} & \tabincell{c}{76.99 \\ \small (\downperf 0.27)} & \tabincell{c}{1964.3 \\ \small (\downperf 18.2)} & \tabincell{c}{54.93 \\ \small (\downperf 4.93)} & \tabincell{c}{47.60 \\ \small (\downperf 10.20)} & \tabincell{c}{82.30 \\ \small (\downperf 5.10)} & \tabincell{c}{53.50 \\ \small (\downperf 8.30)} & \tabincell{c}{89.24 \\ \small (\downperf 4.76)} & \tabincell{c}{58.26 \\ \small (\downperf 35.26)} & \tabincell{c}{75.58 \\ \small (\downperf 5.67)} & \downperf 9.40 \\ \bottomrule
\end{tabular}
\end{adjustbox}
\caption{Results of drop ratio ($\Delta$) when random permutation is applied. We run five experiments.}
\label{main_tab:rq1_main}
\vspace{-1em}
\end{table}}

\paragraph{LLVMs are permutation invariant in terms of visual patch tokens.} From our above results, we empirically verify that each visual patch token from the projector contains localized visual information. Here, we aim to understand how LLVMs systematically process and perceive images based on these visual patch tokens. Given that LLVMs generate answers in an autoregressive manner, we investigate whether they exhibit \textit{order bias} regarding visual patch tokens. To study this, we strongly hypothesize that if LLVMs have \textit{permutation variance}, the performance drop ($\Delta$) will be significant when a random permutation is applied to the visual patch tokens $\mathbf{X}_\text{V}$.

As shown in Table~\ref{main_tab:rq1_main}, the overall performance across most benchmarks declines when the visual patch tokens are randomly shuffled. However, the performance gap between the original and the shuffled (\texttt{Perm.}) versions is not substantial, remaining within a 0–2$\%$ range, for \texttt{LLaVA-1.5} and \texttt{LLaVA-NeXT}. Considering that \texttt{LLaVA-1.5} uses 576 visual tokens, this is an intriguing observation. It suggests that current LLVMs interpret images in a \textit{global} manner, despite each visual patch token containing localized information (see Figure~\ref{main_fig:vis_local_results}), and even though they process both images and text autoregressively. In the case of \texttt{LLaVA-OneVision} which has many visual tokens (729), the avg. performance drop ($\Delta$) is non-trivial. We hypothesize that this global interpretation may result from recent LLVMs being trained via backpropagation, with the loss signal primarily derived from the text output of the \texttt{Assistant:} side. Based on these experiments, we argue that while LLVMs are trained with an autoregressive objective, they internally handle images globally. This observation offers a possible explanation for the success of pixel shuffling~\citep{chen2024far} in achieving both strong performance and efficiency.

\begin{wrapfigure}{r}{0.45\linewidth}
    \centering
    \vspace{-4em}
    \includegraphics[width=\linewidth]{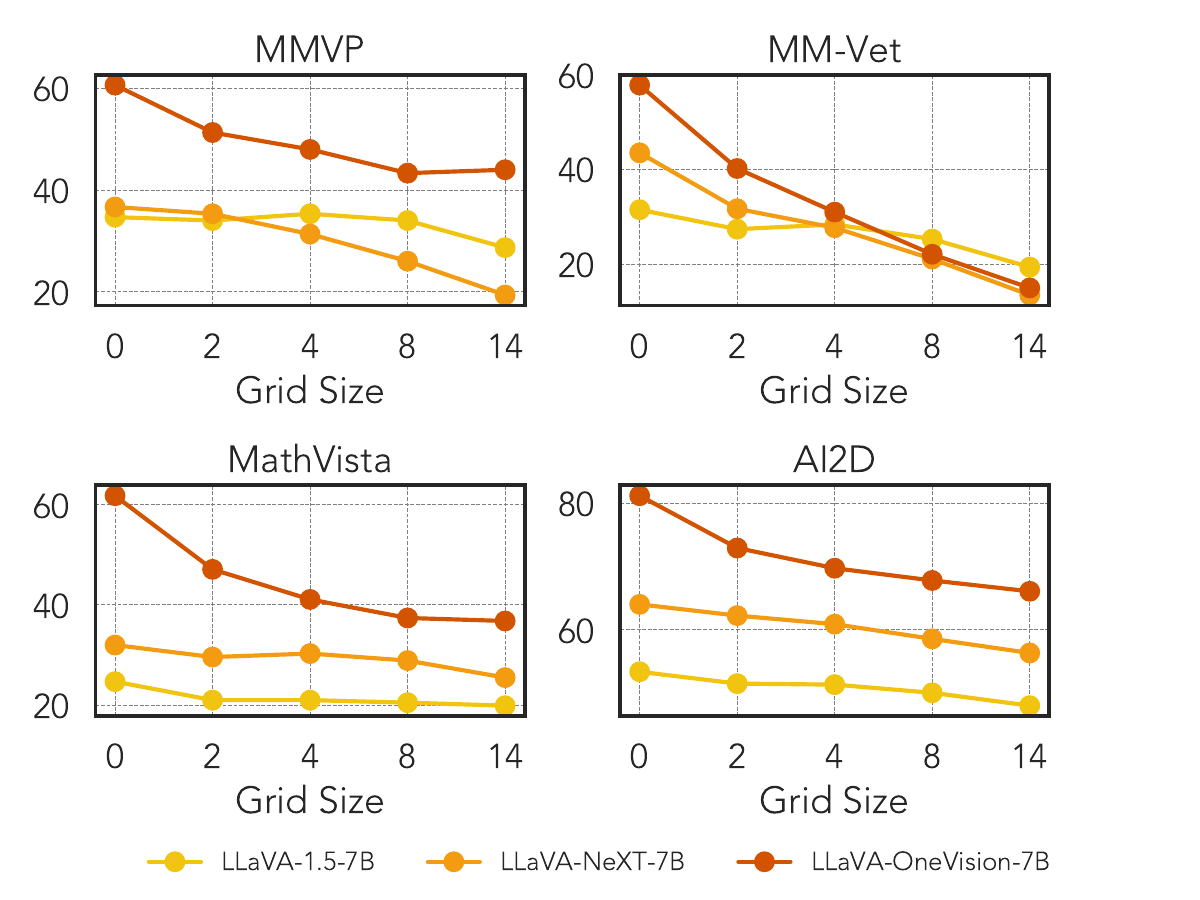}
    \vspace{-2em}
    \caption{
        We present the performance across different grid sizes (2, 4, 8, 14) on the MMVP, MM-Vet, MathVista, and AI2D datasets, using three models: \texttt{LLaVA-1.5}, \texttt{LLaVA-NeXT}, and \texttt{LLaVA-OneVision}.
    }
    \label{main_fig:shuffle_result}
\end{wrapfigure}

\paragraph{LLVMs are sensitive to spatial structures.} Instead of treating visual patch tokens as permutation invariants, we explore how LLVMs behave when the sequence of image patches is permuted. To examine the sensitivity to spatial structure, we randomly shuffle image patches at varying grid sizes (2, 4, 8, 14), as shown in Figure~\ref{main_fig:shuffle_result}. In our experiments, we observe that \texttt{LLaVA-OneVision} is sensitive to spatial structures on the MathVista~\citep{lu2023mathvista} and AI2D~\citep{kembhavi2016diagram} tasks, despite the ViT learning all interactions between image patches. This result contrasts with previous study~\citep{naseer2021intriguing} suggesting that ViT-based vision encoders exhibit high permutation invariance to patch positions than CNN counterparts. We posit that on the MMVP~\cite{tong2024eyes} dataset, which involves perception task, \texttt{LLaVA-OneVision} would also show permutation invariance to randomly shuffled patches, similar to existing work~\citep{naseer2021intriguing} analyzing the ImageNet~\citep{deng2009imagenet} val. dataset. However, unlike ImageNet, the MathVista and AI2D datasets contain more structurally complex images (\eg charts, code screenshots) that are highly sensitive to spatial structure, as the original numerical understanding is significantly disrupted. Interestingly, both \texttt{LLaVA-1.5} and \texttt{LLaVA-NeXT} exhibit insensitivity to spatial structure, particularly in the MathVista dataset, where performance drops were minimal. These results suggest the need for further investigation, which we address in the following sections.

\begin{figure}[tb!]
    \centering
    \begin{minipage}{.38\textwidth}
    \caption{
        We present examples of shuffled images with different grid sizes (2, 4, 8, 14) derived from a MathVista dataset image. As the grid size increases, the chart image becomes more artistically styled.
    }
     \label{fig:patchdrop_vis}
    \end{minipage}%
    \hfill
    \begin{minipage}{.6\textwidth}\vspace{-1em}
    \includegraphics[width=\linewidth]{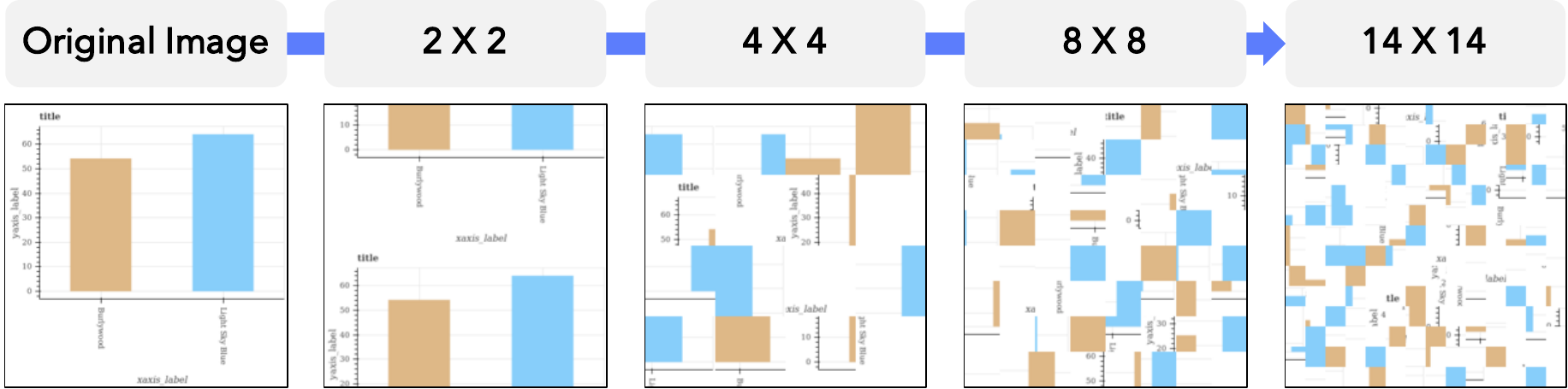}
    \end{minipage}
    \vspace{-1em}
\end{figure}

\subsection{Do LLVMs perceive numerical information well?} \label{sec:rq_synthetic}

Here, we study whether LLVMs truly perceive text-rich images (\eg charts, geometric shapes) that contain highly detailed numerical and shape information. To do this, we construct synthetic datasets for MMVP~\citep{tong2024eyes} and MathVista~\citep{lu2023mathvista}. Specifically, we first generate an image description of a given image using \texttt{LLaVA-OneVision}~\citep{li2024llavaone} with the prompt: ``\textit{Please generate a caption of this image.}''. Next, we generate an image corresponding to the image description leveraging the \texttt{SDXL-Lightning}~\citep{lin2024sdxl} model, ensuring both quality and efficiency. As a result, we get synthesized version (\texttt{Syn.}) to the original version (\texttt{Org.}), illustrated in Figure~\ref{main_fig:occl_syn_examples}.

\begin{wraptable}{r}{0.55\linewidth}
    \vspace{-2em}
    \setlength\tabcolsep{3.2pt}
    \scalebox{0.65}{
    \begin{tabular}{lccccccc}
    \toprule
                  &       &  & \multicolumn{2}{c}{MMVP} & \multicolumn{3}{c}{MathVista} \\ \cmidrule{2-8}
    LLVMs          & Text-rich? & Scale & \texttt{Orig.}  & \texttt{Syn.}  & \texttt{Orig.} & \texttt{Syn.} & Freq. \\ \midrule
    \texttt{CLIP-I} & -      & -    & -               & 0.84           & -              & 0.61       & -    \\ \cdashline{1-8}
    LLaVA-1.5      &   \ding{55}     &   158K   & 34.7            & \tabincell{c}{20.7 \\ \small (\downperf 14.0)}  & 24.7 & \tabincell{c}{22.9 \\ \small (\downperf 1.8)} &  81.0  \\
    LLaVA-NeXT     &   \ding{51}     &   760K   & 36.7            & \tabincell{c}{16.7 \\ \small (\downperf 20.0)}  & 32.0 & \tabincell{c}{27.7 \\ \small (\downperf 4.3)}  & 50.0 \\
    Meteor &    \ding{51}    &   1.1M   & 51.3           & \tabincell{c}{35.3 \\ \small (\downperf 16.0)}  & 52.1 & \tabincell{c}{31.4 \\ \small (\downperf 20.7)} & 9.5  \\ 
    LLaVA-OneVision &    \ding{51}    &   3.1M   & 60.7           & \tabincell{c}{37.3 \\ \small (\downperf 23.3)}  & 61.8 & \tabincell{c}{37.0 \\ \small (\downperf 24.8)} & 12.0  \\ 
    \bottomrule
    \end{tabular}}
    \caption{\small We present the performance on the synthesized versions of the MMVP~\citep{tong2024eyes} and MathVista~\citep{lu2023mathvista} datasets across the models \texttt{LLaVA-1.5}, \texttt{LLaVA-NeXT}, \texttt{Meteor}, and \texttt{LLaVA-OneVision}. Additionally, we provide the scale of the visual instruction training datasets used by each model and specify whether chart, math, and diagram datasets were included. \texttt{CLIP-I} indicates the image similarity using \texttt{CLIP-ViT-L/14}. Freq. denotes the frequency with which the model generates the answer ``1'' in free-form question types.
    }
    \label{main_tab:synthetic_result}
    \vspace{-1em}
\end{wraptable}

\paragraph{In some cases, LLVMs can solve problems without seeing the image.}
Table~\ref{main_tab:synthetic_result} presents the performance comparison on both original and synthesized datasets. For comparison, we evaluate the knowledge-embedded-specific LLVM, \texttt{Meteor 7B}~\citep{lee2024meteor}. Overall, compared to the basic perception task (\ie MMVP), the performance drop in MathVista is not significantly larger across four LLVMs. Given that the generated images show distorted chart and function shapes, with detailed numerical and formula information missing, as shown in Figure~\ref{main_fig:occl_syn_examples} (\texttt{CLIP-I} scores lower than in MMVP), it is surprising that LLVMs are still able to solve math problems requiring advanced cognitive reasoning, without key information. This observation leads us to more in-depth analysis on MathVista dataset.
We analyze how LLVMs solve math problems using synthesized images. In instances where they answer correctly, LLVMs frequently choose ``\textit{No}'' for MCQs and tend to generate ``\textit{1}'' for free-form responses. A deeper analysis reveals that many of these questions ask ``\textit{What is the smallest value?}'', causing the models to select ``1'' using commonsense reasoning, without needing to interpret the image itself. Table~\ref{main_tab:synthetic_result} shows how often the models produce ``1,'' with a noticeable drop in frequency for \texttt{LLaVA-OneVision} and \texttt{Meteor} models. This suggests that these models, likely due to extensive training with million-scale datasets, struggle with ``smallest value'' questions when images are unclear, demonstrating their ability to interpret images effectively.

\begin{wrapfigure}{r}{0.45\linewidth}
    \centering
    \vspace{-2em}
    \includegraphics[width=0.9\linewidth]{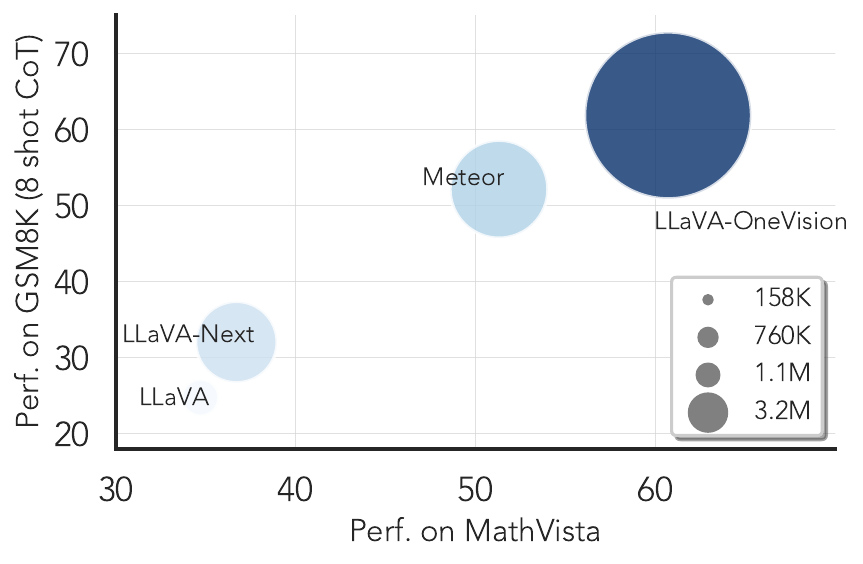}
    \vspace{-1em}
    \caption{
        \small We present performance on the GSM8K dataset using 8-shot Chain-of-Thought prompting. Additionally, we demonstrate that scaling up the instruction-tuning dataset enables LLVMs to solve text-only math reasoning problems more effectively.
    }
    \label{main_fig:gsm_result}
\end{wrapfigure}

\paragraph{Scaling up visual instruction tuning datasets improves text-only mathematical reasoning.} Here, we explore whether enhancing math reasoning in a visual context can improve standard text-only math reasoning. We evaluate four LLVMs on the GSM8K~\citep{cobbe2021training} dataset in an 8-shot setting using Chain-of-Thought (CoT) prompting~\citep{wei2022chain}. As shown in Figure~\ref{main_fig:gsm_result}, we observe that models performing well in visual math contexts also achieve strong performance on GSM8K. Moreover, as the scale of the dataset used for training increases, so does model performance. This suggests that using high-quality, large-scale datasets (\eg rationale-style datasets, as used in \texttt{Meteor}) is beneficial, and that there is compatibility between visual math and text-only math reasoning, aligning with the data-centric AI perspective~\citep{xu2023demystifying,zhou2024lima}.

\begin{figure}[tb!]
    \centering
    \begin{minipage}{.38\textwidth}
    \caption{
        \small We present examples of images (left) synthesized by \texttt{SDXL-Lightning} and (right) occluded using three methods: \texttt{Random}, \texttt{Salient}, and \texttt{Non-Salient}. The original images are from the MathVista and MME datasets. Occluded areas are marked in black to indicate zero pixel values.
    }
     \label{main_fig:occl_syn_examples}
    \end{minipage}%
    \hfill
    \begin{minipage}{.6\textwidth}\vspace{-1em}
    \includegraphics[width=\linewidth]{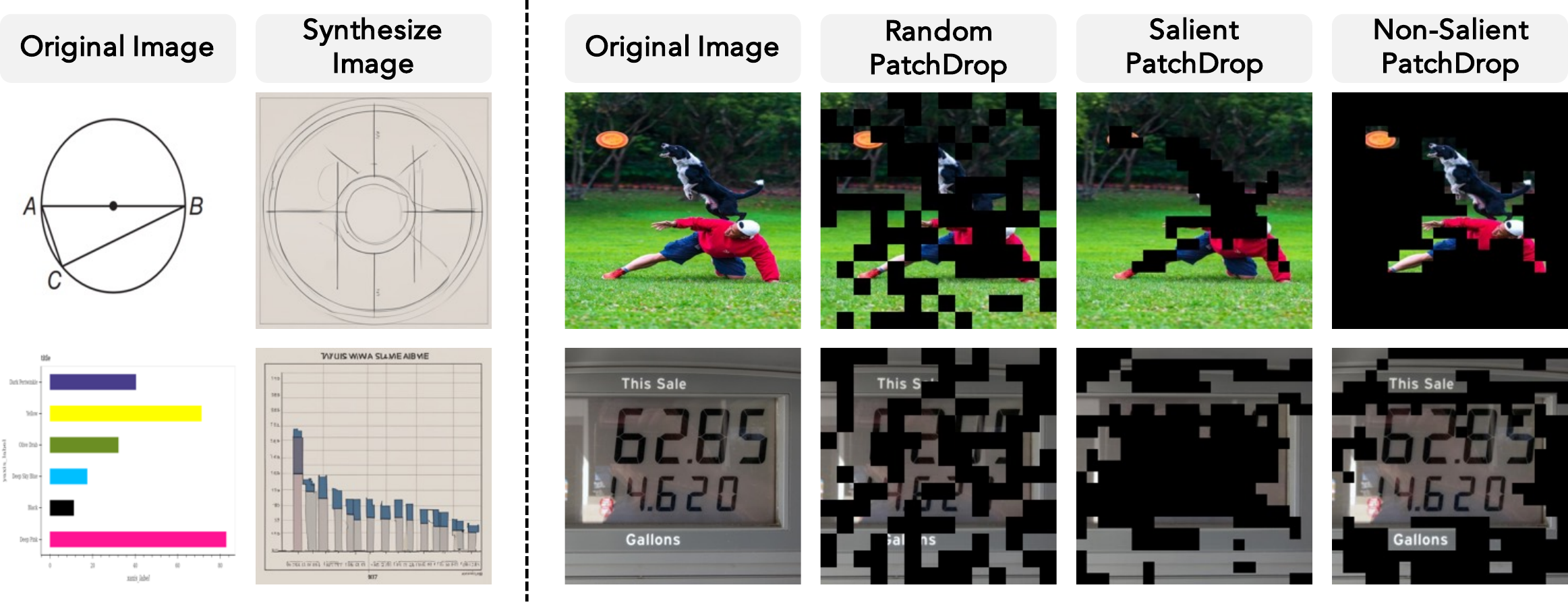}
    \end{minipage}
    \vspace{-1em}
\end{figure}

\subsection{Are LLVMs Robust to Occlusions?}

Existing studies~\citep{naseer2021intriguing,vishniakov2023convnet} have demonstrated that ViT models exhibit a remarkable degree of robustness to occlusions, such as patch dropping, than CNN counterparts. Since most LVLMs utilize \texttt{CLIP ViT-L} as their vision encoder, we aim to explore whether this robustness transfers to LVLMs in scenarios involving occluded images. Following the simple masking method presented in prior work~\citep{naseer2021intriguing}, we manipulate the input image $I = \{x_i\}_{i=1}^N$, where $N$ represents the number of patches. Specifically, we mask out $N'$ patches (where $N' < N$) by setting the pixel values of these patches to zero, creating an occluded image, denoted as $\tilde{I}$. We then apply three distinct occlusion methods to the image $I$: (1) \texttt{Random PatchDrop}: A subset of $N'$ patches is randomly selected and dropped from the image, effectively simulating random occlusion; (2) \texttt{Salient PatchDrop}: We strategically select and drop salient patches by leveraging the self-supervised ViT model \texttt{dino-small}~\citep{caron2021emerging}; (3) \texttt{Non-Salient PatchDrop}: In this case, we drop non-salient, background patches, retaining only the salient information. This method also utilizes \texttt{dino-small}, following a similar approach to the \texttt{Salient PatchDrop} but focusing on removing the background regions. Figure~\ref{main_fig:occl_syn_examples} presents example images with different occlusion methods applied.

\begin{figure}[tb!]
    \centering
    \begin{minipage}{.7\textwidth}
    \includegraphics[width=\linewidth]{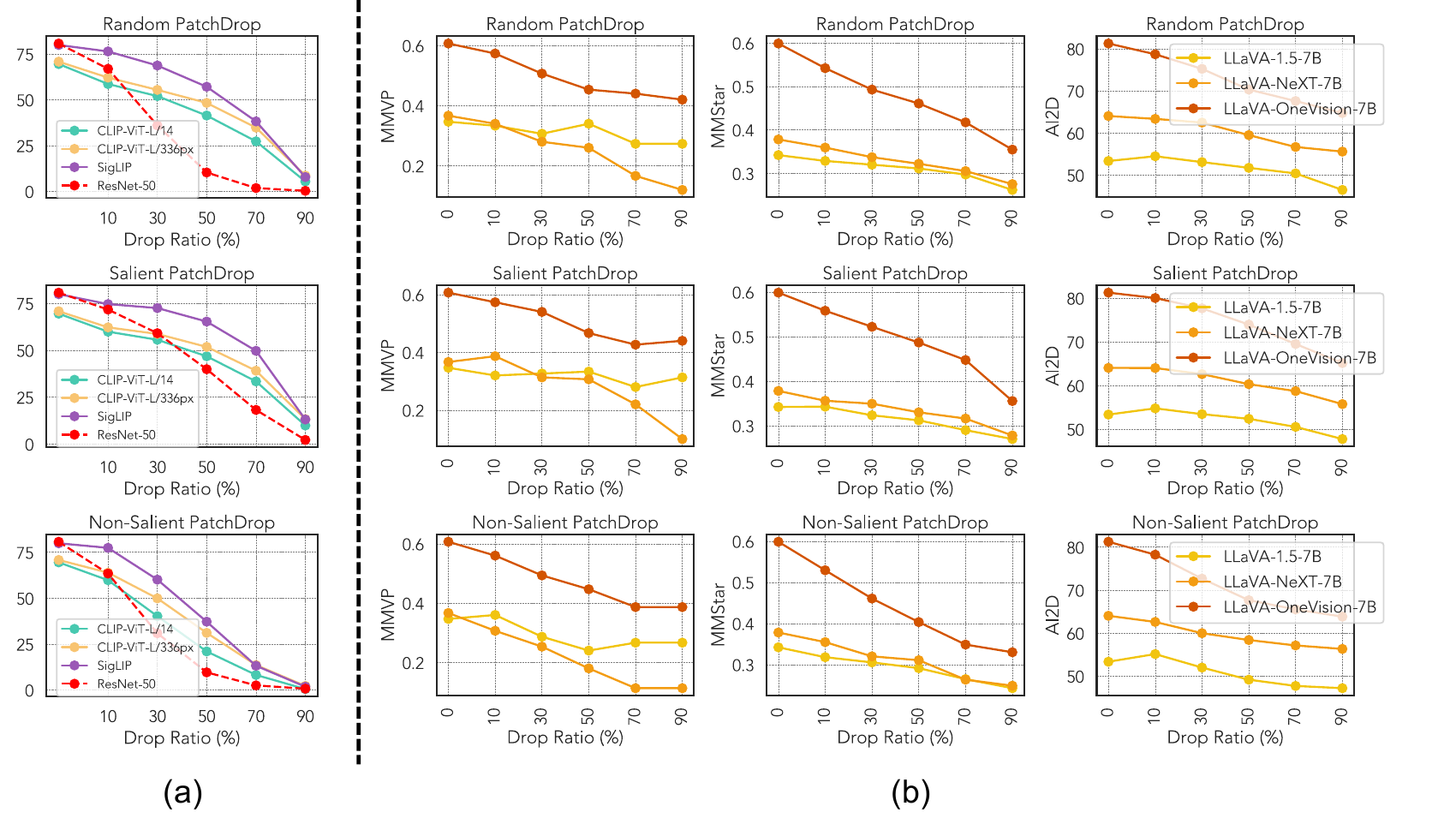}
    \end{minipage}%
    \hfill
    \begin{minipage}{.28\textwidth}
    \caption{
        We present robustness performance under occlusion conditions. (a) ViT variant vision encoders demonstrate greater robustness to occlusion compared to \texttt{ResNet-50}. (b) LLVMs also show robustness to occlusion, benefiting from the use of ViT encoders.
    }
     \label{main_fig:occlusion_result}
    \end{minipage}
    \vspace{-1em}
\end{figure}

\paragraph{LLVMs are robust against occlusion.}

Before evaluating LLVMs on occluded images, we first verify whether ViT-based encoders are more robust than their CNN counterparts in this scenario. To do this, we assess several ViT variants and \texttt{ResNet-50}~\citep{he2016deep} on occluded ImageNet~\citep{krizhevsky2009learning} images, applying the same masking process as mentioned above. As shown in Figure~\ref{main_fig:occlusion_result} (left), compared to \texttt{ResNet-50}, ViT variants demonstrate greater robustness in occlusion scenarios, consistent with findings from the prior study~\citep{naseer2021intriguing}. Due to this robustness, LLVMs also exhibit relatively strong performance under occlusion. This result is surprising given that in the AI2D dataset, which contains text-rich diagram images, 50\%–70\% of the patches are missing, yet LLVMs can still provide correct answers to some extent. This may be because AI2D involves selecting one answer from multiple options, suggesting the possibility of selection bias~\citep{zheng2023large}, a significant issue that we leave for future work.

\subsection{Do LLVMs preserve cross-modal alignment?}

In the \textit{de facto} structure of LLVMs, a projector $f_\text{P}$ enables LLMs to perceive and understand images by transforming visual representations into the LLM’s representation space. While a recent work~\citep{mckinzie2024mm1} suggests that the type of projector has minimal impact on performance, other studies~\citep{zhai2024investigating,verma2024cross} have argued that the projector have limitations in preserving cross-modal understanding and issues such as catastrophic forgetting. In this work, we investigate (1) how effectively a trained projector maintains its \textit{visual recognition} capability relative to the LLVM’s original vision encoders (\eg \texttt{CLIP-ViT-14} for \texttt{LLaVA-NeXT}), and (2) how well a trained projector preserves cross-modal alignment, based on the \textit{platonic representation hypothesis}~\citep{huh2024platonic}, compared to representation expressivity without alignment learning.

{\renewcommand{\arraystretch}{1.35}
\begin{table}[!t]
\centering
\begin{adjustbox}{width=\linewidth}
\begin{tabular}{lcccccccc}
\toprule 
Models          & Caltech100 & CIFAR-100 & Food101 & Pets & Country211 & EuroSAT  & AirCraft & Avg.  \\ \midrule
CLIP ViT 336    & 85.2       & 76.9      & 92.9    & 93.0 & 31.7       & 59.0    & 33.2     & 67.4   \\
\grayrow LLaVA-1.5           & 44.4 \small (\downperf 40.8) & 50.8  \small (\downperf 26.1)    & 57.9 \small (\downperf 35.0)  & 73.0  \small (\downperf 20.0)                          & 12.2 \small (\downperf 19.5)     & 11.8 \small (\downperf 47.2)  & 17.6 \small (\downperf 15.6) & 38.2 \small (\downperf 29.2) \\
CLIP ViT 14     & 85.2       & 76.3      & 92.4    & 93.2 & 27.5       & 57.7    & 32.9     & 66.5 \\
\grayrow  LLaVA-NeXT      & 57.2 \small (\downperf 28.0)                                   & 56.3 \small (\downperf 20.0)                    & 64.5 \small (\downperf 27.9)                        & 76.1  \small (\downperf 17.1)    & 14.6 \small (\downperf 12.9)  & 23.8  \small (\downperf 33.9)              & 18.3 \small (\downperf 14.6)  & 44.4  \small (\downperf 22.1)   \\ \bottomrule
\end{tabular}
\end{adjustbox}
\caption{We report the Top-1 accuracy ($\%$) of LLVMs and their corresponding vision encoder models on 1K subsampled datasets from Caltech100, CIFAR-100, Food101, Pets, Country211, EuroSAT, and AirCraft.}
\label{main_tab:zero_image_clf}
\vspace{-1em}
\end{table}}

\paragraph{LLVMs struggle to preserve the original visual understanding capability.} Ideally, after alignment and visual instruction tuning, LLVMs should retain the visual perception abilities of their original vision encoders, allowing them to understand and classify images effectively. To assess this, we evaluate LLVMs on zero-shot image classification tasks using widely adopted datasets such as Caltech100~\citep{higgins2017beta}, Food101~\citep{bossard2014food}, CIFAR-100~\citep{krizhevsky2009learning}, Pets~\citep{parkhi2012cats}, Country211~\footnote{\url{https://github.com/openai/CLIP/blob/main/data/country211.md}}, EuroSAT~\citep{helber2019eurosat}, and AirCraft~\citep{maji2013fine}. Following the method in previous work~\citep{zhai2024investigating}, we use ChatGPT (\texttt{gpt-3.5-turbo})~\citep{chatgpt} to extract a single label with the use of prompt: \textit{Is this prediction correct?}.
As shown in Table~\ref{main_tab:zero_image_clf}, the performance of LLVMs significantly degrades across all datasets compared to their vision encoders, suggesting that LLVMs do not fully retain the perception capabilities of their original vision encoders. This may be due to: (1) LLVMs being trained to solve complex tasks (\eg chart or math reasoning) with the use of instruction, which may cause them to lose basic perception abilities (\eg recognizing simple objects), a phenomenon known as \textit{catastrophic forgetting}~\citep{zhai2024investigating}~\footnote{On the CLEVR/Count~\citep{johnson2017clevr} dataset, we observed a 16.6\% performance improvement in the \texttt{LLaVA-NeXT} model compared to the previous vision encoder (\ie \texttt{CLIP-ViT-L/14}.)}, and (2) the vision encoder's relatively small parameter size (307M for \texttt{CLIP ViT-L/336px}) compared to the LLM (7B for \texttt{Vicuna-1.5}), which could result in a loss of visual perception capability during projection, as the more powerful LLM dominates.

\begin{wrapfigure}{r}{0.45\linewidth}
    \centering
    \vspace{-2em}
    \includegraphics[width=\linewidth]{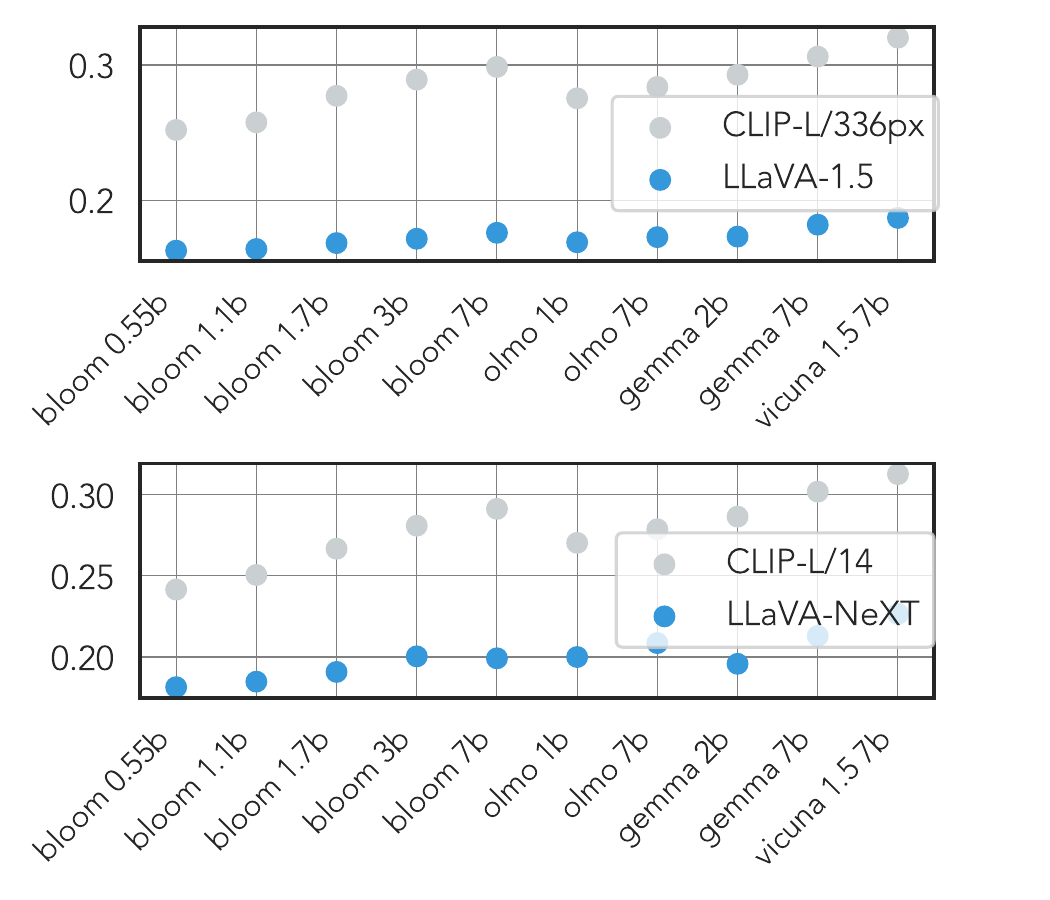}
    \vspace{-2em}
    \caption{
        We present how alignment preservation changes (\texttt{CLIP} $\rightarrow$ \texttt{LLaVA}) in the representation space across various LLM families, \texttt{BLOOM}~\citep{le2023bloom}, \texttt{OLMo}~\citep{groeneveld2024olmo}, \texttt{Gemma}~\citep{team2024gemma}, \texttt{Vicuna}~\citep{vicuna2023}, with different parameter sizes on the DOCCI dataset.
    }
    \label{main_fig:platonic}
\end{wrapfigure}

\paragraph{LLVMs lose the ability to understand and interpret shared world concepts.} Beyond visual recognition capabilities, we analyze cross-modal alignment based on the \textit{platonic representation hypothesis}~\citep{huh2024platonic}, which argues that neural networks, despite being trained on different objectives, data, and modalities, should converge to a shared statistical model of reality in their representation spaces. To measure representation similarity between two modalities, the original authors of this hypothesis use mutual nearest-neighbor alignment metrics, a type of kernel-alignment metric. In our work, we assess how much alignment is lost after visual instruction tuning by applying this metric within the context of the platonic representation hypothesis. We evaluate 10 LLMs and measure alignment between these LLMs and vision encoders (LLVMs) using the DOCCI~\citep{onoe2024docci} dataset which contains long image descriptions requiring localized scene understanding. 
As shown in Figure~\ref{main_fig:platonic}, after visual instruction tuning, both \texttt{LLaVA-1.5} and \texttt{LLaVA-NeXT} show degraded alignment performance with respect to representations compared to their original vision encoder. This suggests doubts about the actual role of the projector in causing the degradation in alignment preservation. From this observation, we speculate that current LLVMs are trained on a variety of datasets to achieve generalization (\ie multi-task learning). However, during visual instruction tuning, the models might overemphasize capabilities requiring complex cognition while potentially reducing representations related to other tasks, such as localized scene understanding (\ie DOCCI). This results in a lower alignment score and catastrophic forgetting, as shown in Table~\ref{main_tab:zero_image_clf}. For future work, one potential direction is to develop a localized enhanced alignment module similar to HoneyBee~\citep{cha2024honeybee}.

\subsection{Model behavior: Which modality and layers are more important?}

Here, we conduct an in-depth analysis of model behavior to assess the importance of either a layer or a visual token when performing downstream tasks. We hypothesize that if adding arbitrary noise to a specific component—either a layer block or a visual token—results in a significant drop in model performance, then that component is crucial and actively involved in the model's reasoning process. To quantify this, we define an \textit{importance} score ($\mathcal{I}$) inspired by the concept of ``\textit{sharpness of minima}''~\citep{keskar2016large,lee2024fp8}. This concept aims to identify flat minima, which promote stable training and better generalization, by measuring the sensitivity of the training function around a local minimum. In our work, we adapt this concept for the inference stage.

\begin{definition}[Importance Score] 
    Let $x_t \in \mathbb{R}^d$ be the $d$-dimensional input embedding for a target subject $t$. For $x_t$, we define the constraint candidate set $\mathcal{C}_t$ for $t$ as: 
    \begin{equation} 
        \mathcal{C}_t = \{z_t \in \mathbb{R}^d : -\epsilon + |x_t| \leq z_t \leq \epsilon + |x_t|\}, \quad \epsilon \sim \text{Uniform}(-1, 1), 
    \end{equation} 
    where $z_t$ is a noise vector. The importance score $\mathcal{I}$ for target $t$ is then defined as: 
    \begin{equation} 
        \mathcal{I}_t := \frac{f(x_t) - \max_{z_t \in \mathcal{C}_t}f(x_t + z_t)}{f(x_t)} \times 100. 
    \end{equation} 
\end{definition}

Note that while the concept of ``sharpness of minima'' was originally used to find flat minima during training by defining a square-bound constraint set, this is feasible because the model is trained via stochastic gradient descent (SGD), which indirectly allows the evaluation of all noise values $z$ in the constraint set $C$. However, since our experiment focuses on downstream task performance during inference, we adapt this concept by sampling $K$ noise vectors, $\{z_t^1, z_t^2, \dots, z_t^K \} \sim \mathcal{C}_t$, with different random seeds. For computational efficiency, we set $K=10$.

\begin{wrapfigure}{r}{0.3\linewidth}
    \centering
    \vspace{-3em}
    \includegraphics[width=\linewidth]{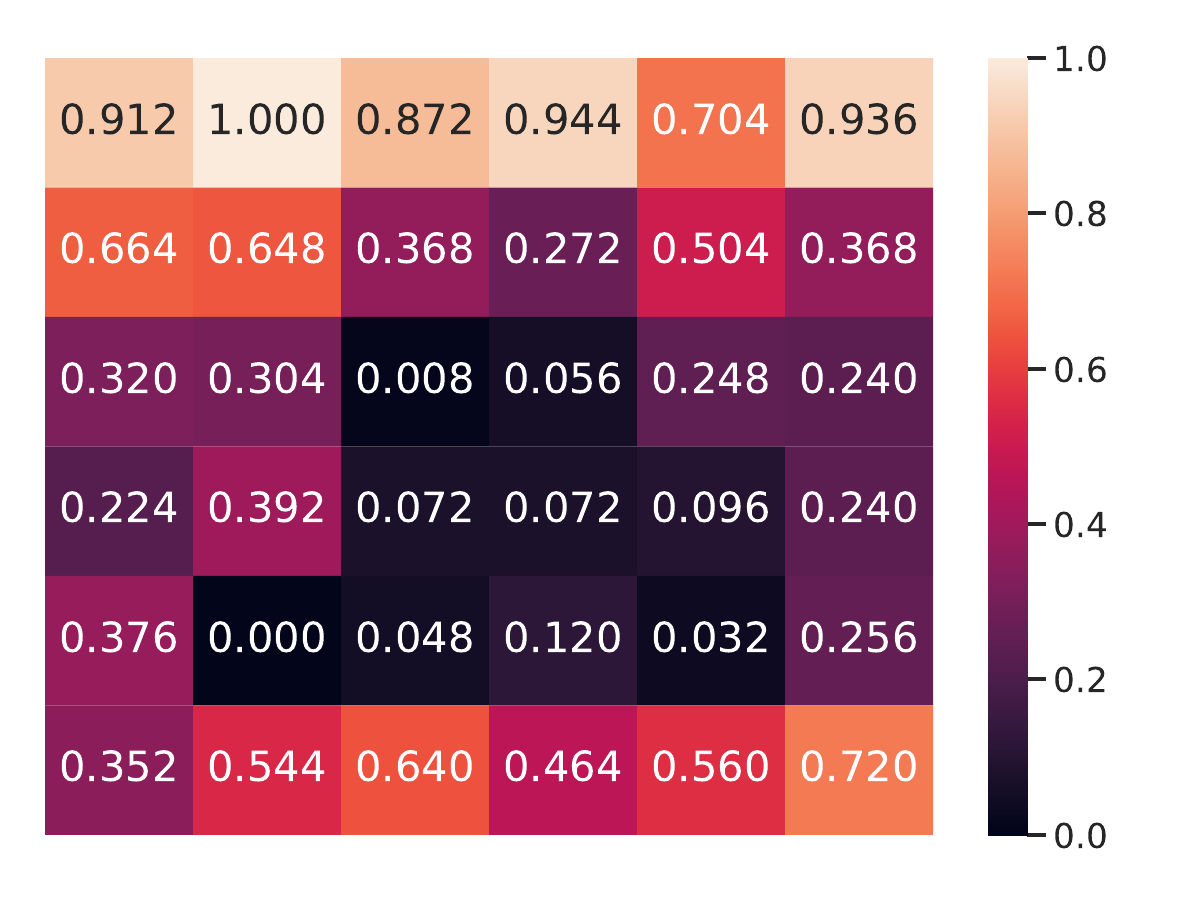}
    \vspace{-2em}
    \caption{
        We report the degree of utility of group-wise visual tokens for LLaVA 1.5 7B on the MM-Vet dataset. Darker regions indicate that the LLVM relies heavily on information from those specific group parts.
    }
    \vspace{-1em}
    \label{main_fig:vis_tok_importance}
\end{wrapfigure}

\paragraph{LLVMs strongly focus on the center of the image.} To assess the extent to which LLVM utilizes visual token information, we add a noise vector $z_t$ to each visual token information based on the importance score ($\mathcal{I}$). However, performing computations on each individual visual token is computationally intensive, especially for models such as \texttt{LLaVA-1.5-7B} that process 576 visual tokens arranged in a $24 \times 24$ grid of patches. To accelerate computation and reduce complexity, we adopt the same process as \ref{sec:rq1}. As shown in Figure~\ref{main_fig:vis_tok_importance}, \texttt{LLaVA-1.5-7B} places strong emphasis on the central part of the images in the MM-Vet dataset, while the edge regions have minimal influence on the final performance compared to the central visual tokens. This result suggests that not all visual tokens are necessary, aligning with recent works~\citep{cha2024honeybee,xue2024xgen} that reduce redundant visual tokens in the projector to enhance efficiency.

\begin{wrapfigure}{r}{0.5\linewidth}
    \centering
    \vspace{-3em}
    \includegraphics[width=\linewidth]{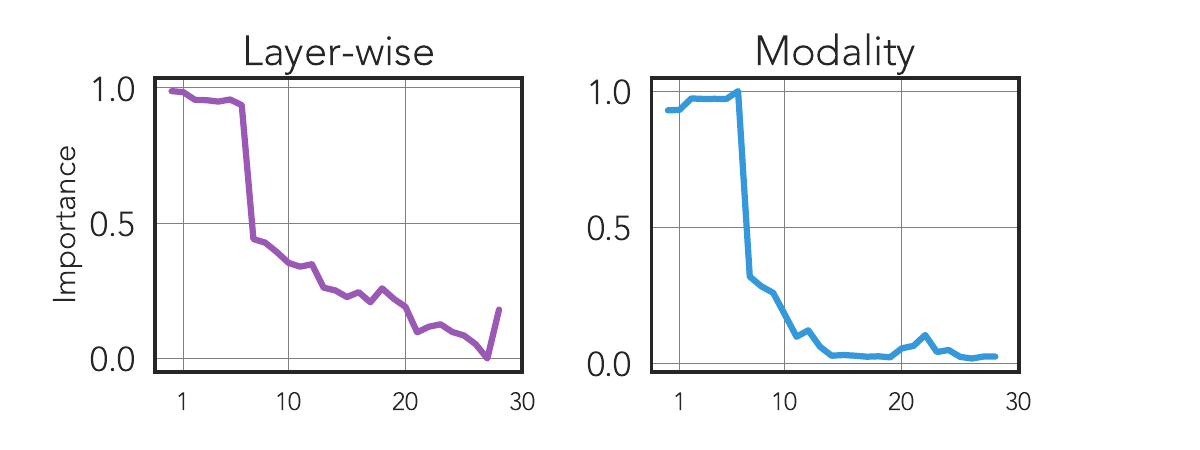}
    \caption{
        We present the results of (left) layer-wise importance and (right) modality importance within the layers.
    }
    \vspace{-1em}
    \label{main_fig:layer_importance}
\end{wrapfigure}

\paragraph{Lower block layers in LLVMs are more important.} Figure~\ref{main_fig:layer_importance} (left) shows that the lower layers ($<$ 6) play a crucial role in handling the integrated capabilities required for tasks in the MMStar dataset. This suggests that these layers contain more beneficial representations for perception and cognition. This finding aligns with recent work on LLVMs, specifically TroL~\citep{lee2024trol}, which introduces the concept of ``\textit{layer traversal}.'' This technique revisits layer representations, resulting in a highly generalizable model despite its small size (1.8B parameters). In their paper (Figure 6), the traversal pattern is more pronounced in the lower layers, which is consistent with our findings. Therefore, we believe our results may provide insights into why traversing layers leads to improved generalization.

\vspace{-1em}

\paragraph{Textual modality is more important than visual modality.} In addition to layer-wise importance, we measure \textit{modality importance} using the score $\frac{\mathcal{I}_I}{\mathcal{I}_T}$, which calculates the relative importance of textual and image modalities. Specifically, to obtain the image modality importance score $\mathcal{I}_I$, we feed noise vectors to the positions corresponding to image tokens (\eg 576 tokens in \texttt{LLaVA-1.5}), and vice versa for text modality $\mathcal{I}_T$. As shown in Figure~\ref{main_fig:layer_importance} (right), until the lower layers ($< 8$), the image modality is more important than the textual modality. However, as layers progress, we observe that the textual modality becomes increasingly important, likely because generating responses requires a stronger focus on text with the perspective of autoregressive modeling. This suggests that LLVMs initially focus on global image perception (\cref{sec:rq1}), and by the middle layers, they have processed the image and shifted toward solving complex user queries to generate coherent answers. Similarly, in TroL, layer traversal occurs more actively in the lower layers, which we interpret as the model attempting to better comprehend the image when it fails to do so in a single pass, enabling it to solve complex reasoning tasks more effectively. These findings highlight the value of strong visual perception, which may explain the success of models utilizing large visual tokens~\citep{wang2024qwen2,li2024llavaone} or high-resolution image processing~\citep{li2024llavaone}.

\section{Discussions} \label{main_sec:discussion}

\paragraph{Building more interactive evaluation benchmarks.} As mentioned in \cref{sec:rq_synthetic}, LLVM can effectively solve problems even without seeing the input image. However, current evaluation benchmarks are designed for single-turn interactions and lack applicability to real-world, interactive scenarios. For example, in standard OCR tasks, we typically assess whether the LLVM correctly transcribes text from an image. But consider a practical situation: you're traveling in a foreign country and visiting a local restaurant. Translating the menu is challenging, and while an application with strong OCR capabilities would be helpful, this is only the first step. When ordering, the LLVM should not only recognize the menu items but also understand the user's preferences — what they like or dislike — by incorporating knowledge of their persona~\citep{lee2022personachatgen}. Therefore, future benchmarks should be more interactive and socially grounded~\citep{zhou2023sotopia}, extending beyond multiple-choice, binary, or non-interactive free-form tasks. These benchmarks should involve multi-turn interactions and be based on the user's preferences~\citep{lee2024aligning} or persona in long-term social interactions~\citep{jang2023conversation,lee2024stark}.

\vspace{-1em}

\paragraph{A new paradigm enhancing cross-modal alignment.} Current LLVMs have widely adopted the \textit{model-stitching} structure, which demonstrates impressive capabilities on tasks requiring higher-level cognitive reasoning. However, they exhibit significantly degraded performance in zero-shot image classification tasks (Table~\ref{main_tab:zero_image_clf}). Additionally, they cannot effectively preserve alignment (Figure~\ref{main_fig:platonic}) in terms of relatively simple perception when compared to text-rich images (\eg charts, mathematical expressions). Recently, although recent studies~\citep{li2024llavaone,lee2024meteor} has been extensively scaling up model sizes with larger datasets to achieve higher performance on increasingly difficult tasks — which we believe is the correct direction - we think it is necessary to deeply consider innovative model architectures (\eg layer of traversal~\citep{lee2024trol}, hidden dimension expansion~\citep{lee2024phantom}) to enhance cross-modal alignment at least once. For example, in recent unified architectures~\citep{xie2024show,sharegpt4o}, enabling LLMs to generate images is akin to how drawing can be substantially more challenging for humans than simply viewing a picture. This is because drawing requires a comprehensive and simultaneous understanding of complex world concepts such as relationships between objects, lighting, perspective, and more. Therefore, by projecting visual imagination abilities~\citep{lu2022imagination,lee2023large} onto LLVMs to enable them to generate images, it might significantly help in better preserving cross-modal alignment.

\section{Conclusion}

In this paper, we systemically reveals intriguing properties of LLVMs with respect to \textit{permutation invariance}, \textit{robustness}, \textit{alignment preserving}, and \textit{importance} under various image settings such as occlusion and synthesized images. We hope these findings will assist academic researchers and ML developers in advancing the next frontier of LLMs by providing a foundational basis for future model design choices.

\bibliography{iclr2025_conference}
\bibliographystyle{iclr2025_conference}

\appendix

\section{Description of Evaluation Benchmarks} \label{supp_sec:benchmark_description}

\begin{itemize}
\item{\textbf{MM-Vet}~\citep{yu2023mm}} dataset is a benchmark designed to evaluate large vision-language models (LVLMs) across six core vision-language (VL) capabilities: recognition, knowledge, optical character recognition (OCR), spatial awareness, language generation, and mathematical reasoning. The dataset includes open-ended, real-world questions based on image-text pairs, requiring models to integrate multiple capabilities to solve complex tasks. MM-Vet benchmark consists of 200 images paired with 218 open-ended questions.

\item{\textbf{Q-Bench}~\citep{wu2023q}} evaluates the capabilities of  large vision-language models in three main areas related to low-level vision tasks. These tasks focus on evaluating how well LVLMs can perform basic low-level perception tasks that are traditionally associated with human visual perception. In the Q-Bench dataset, the questions are of three types: Yes-or-No, What, and How.

\begin{itemize}
\item \textbf{Low-Level Visual Perception}: Assesses how accurately LVLMs can answer questions about low-level image attributes (e.g., clarity, color, distortion). LLVisionQA dataset includes 2,990 images, each with a corresponding question about low-level features.

\item \textbf{Low-Level Visual Description}: Evaluates the ability of LVLMs to describe images. LLDescribe dataset has 499 images with expert-labeled descriptions averaging 58 words each. LVLMs are compared against these to assess completeness, preciseness, and relevance.

\item \textbf{Visual Quality Assessment}: Evaluates LVLMs' ability to predict quantifiable quality scores for images by assessing how well they align with human-rated mean opinion scores (MOS) on low-level visual appearances, using 81,284 samples. 
\end{itemize}

\item{\textbf{SQA-IMG}~\citep{lu2022learn}} is a portion of the Science Question Answering (SQA) dataset that contains questions from a wide range of scientific domains, each paired with corresponding image contexts. The dataset includes 10,332 examples of multimodal multiple-choice questions, along with lectures and explanations that detail the reasoning behind the correct answers.

\item{\textbf{ChartQA}~\citep{masry2022chartqa}} dataset is a benchmark designed to test AI models on their ability to perform question-answering tasks over various types of charts. It focuses specifically on questions requiring complex reasoning, such as visual and logical interpretation, going beyond simpler template-based datasets. ChartQA includes 9,608 human-authored open-ended questions as well as 23,111 questions that are automatically generated from chart summaries.

\item{\textbf{SEED-IMG}~\citep{li2023seed}}, a subset of SEED-Bench, focuses on evaluating spatial comprehension of images by testing models on various dimensions like scene understanding, object identification, and spatial relationships. In terms of scale, the dataset includes 19,000 multiple-choice questions that evaluate both image and video comprehension, covering 12 evaluation dimensions such as scene understanding, instance identity, spatial relations, and action recognition.

\item{\textbf{MME}~\citep{fu2023mme}} evaluates both perception and cognition abilities of LVLMs. It features 14 subtasks, including recognition tasks (such as object existence, count, position, color) and reasoning tasks (such as commonsense reasoning, numerical calculation, text translation, and code reasoning). MME uses manually created instruction-answer pairs, ensuring no overlap with public datasets. MME uses "yes/no" responses for quantitative evaluations.

\item{\textbf{MathVista}~\citep{lu2023mathvista}} is a benchmark designed to evaluate the mathematical reasoning capabilities of foundation models in visual contexts. It integrates challenges from diverse mathematical and visual tasks, with a focus on fine-grained, deep visual understanding and compositional reasoning. MathVista consists of 6,141 examples including 3,392 multiple-choice questions and 2,749 free-form questions derived from 28 existing multimodal datasets and 3 newly created datasets: IQTest, FunctionQA, and PaperQA. 

\item{\textbf{LLaVA-W}~\citep{liu2024visual}} is a challenging evaluation benchmark created to assess the generalization and instruction-following capabilities LVLMs in complex, real-world situations. It consists of 24 images and 60 questions, including diverse scenes like indoor environments, outdoor settings, memes, paintings, and sketches. Each image is associated with a highly detailed and manually curated description, and the questions focus on extracting intricate details and reasoning about the visual content. LLaVA-W involves a variety of tasks, including detailed descriptions, conversational answers, and complex reasoning.

\item{\textbf{MMStar}~\citep{chen2024we}} is a vision-dependent multimodal benchmark designed to evaluate the multimodal capabilities of LVLMs. It addresses two main issues identified in previous benchmarks: the reliance on textual information without visual input and data leakage during training. MMStar is composed of 1,500 samples carefully selected 
to ensure that visual content is necessary for solving each problem. MMStar evaluates six core capabilities across 18 detailed axes, which include tasks like image perception and logical reasoning. MMStar uses multiple-choice as the primary answer type.

\item{\textbf{MMVP}~\citep{tong2024eyes}} evaluates the visual grounding capabilities of large vision-language models by identifying scenarios where they fail to distinguish simple visual patterns in images. These patterns include aspects like orientation, counting, viewpoint, and relational context. The benchmark is constructed using 150 pairs of images, resulting in 300 multiple-choice questions.

\end{itemize}

\section{Description of Evaluation LVLMs} \label{supp_sec:lvlm_description}
\begin{itemize}

\item{\textbf{LLaVA-1.5}~\citep{liu2024improved}} incorporates academic task-oriented datasets to enhance performance in VQA tasks and features an MLP vision-language connector, which improves upon the original linear layer utilized in LLaVA~\citep{liu2024visual}. It uses CLIP ViT-L/14~\citep{radford2021learning} with a 336px resolution as its vision encoder, resulting in a total of $(336/14)^2$ = 576 visual tokens. LLaVA-1.5 is built on Vicuna with either 7B or 13B parameters. The training dataset includes 558K samples for pre-training and 665K for fine-tuning, totaling 1.2M image-text pairs from publicly available datasets

\item{\textbf{LLaVA-NeXT}~\citep{liu2024llava}} (also known as LLaVA-1.6) enhances visual reasoning, OCR, and world knowledge, offering four times higher image resolution (up to 1344x336) and improved performance in visual conversations. Its architecture includes a CLIP ViT-L/14 as a vision encoder, paired with Vicuna models ranging from 7B to 34B as a backbone language model. It utilizes 1.3M visual instruction tuning data samples for training, maintaining efficiency with approximately one day of training on 32 A100 GPUs. The architecture's high resolution and dynamic grid scheme improve detailed image processing capabilities.

\item{\textbf{LLaVA-OneVision}~\citep{li2024llava}} is a LVLM designed for task transfer across single-image, multi-image, and video scenarios, with strong capabilities in video understanding through image-to-video task transfer. Its architecture consists of a Qwen2 language model~\citep{yang2024qwen2} with 8B to 72B parameters, and the SigLIP vision encoder~\citep{zhai2023sigmoid}, which processes images at a base resolution of 384x384, producing 729 visual tokens. The model employs a 2-layer MLP projector. The training utilized 3.2M single-image data samples and 1.6M multi-modal data samples, focusing on high-quality visual instruction tuning data to enhance its multimodal capabilities.

\item{\textbf{Meteor}~\citep{lee2024meteor}} is a large vision-language model that uniquely embeds multifaceted rationales using a Mamba-based architecture~\citep{gu2023mamba}, enabling efficient processing of lengthy rationales to enhance its vision-language understanding. This approach allows Meteor to achieve superior performance without scaling up model size or employing additional vision encoders. Its architecture includes a CLIP-L/14 vision encoder with an image resolution of 490x490, comprising 428M parameters, and InternLM2-7B~\citep{cai2024internlm2} as a foundational LLM. Meteor was trained on 2.1M question-answer pairs, with 1.1M curated triples.

\item{\textbf{TroL}~\citep{lee2024trol}} uses a unique characteristic called layer traversing, which reuses layers in a token-wise manner, allowing it to simulate retracing the answering process without physically adding more layers, making it efficient despite smaller model sizes. TroL uses CLIP-L and InternViT as vision encoders, containing 428M and 300M parameters, respectively, and supports 24 layers. The image resolution is adjusted using MLPs in the vision projector. For its foundational LLM, TroL utilizes Phi-3-mini with 3.8B parameters and InternLM2 with 1.8B and 7B parameters. The training dataset comprises 2.3M visual instruction tuning samples.

\end{itemize}

\end{document}